# Can Machines Think in Radio Language?

Yujian Li

College of Computer Science, Beijing University of Technology, Beijing 100124, China

Email: liyujian@bjut.edu.cn

**Abstract**: *People can think in auditory, visual and tactile forms of language, so can machines principally. But is it possible for them to think in radio language? According to a first principle presented for general intelligence, i.e. the principle of language's relativity, the answer may give an exceptional solution for robot astronauts to talk with each other in space exploration.*

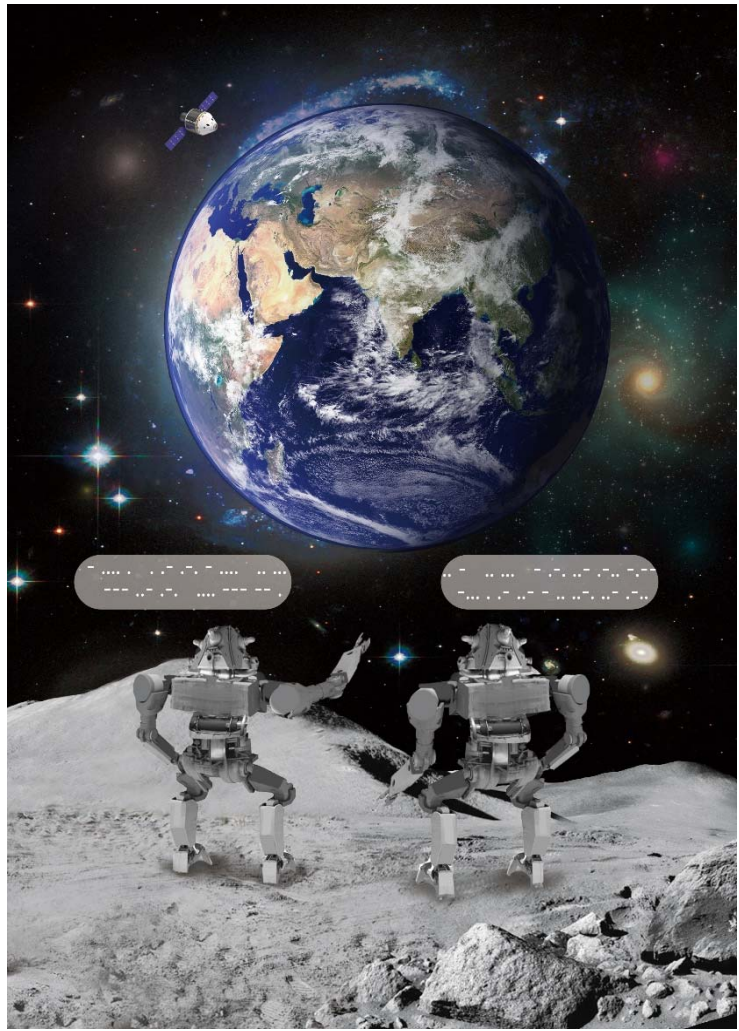

Figure 1. Two robonauts are talking about the earth in a Morse-code radio language on the moon. One says "THE EARTH IS OUR HOME", the other "IT IS TRULY BEAUTIFUL". This is a potential application of Morse code to artificial intelligence in the future, beyond classical transmission of text information between people.

In computer science, one of the biggest unsolved problems is to develop intelligent machines. Since a seminal paper by Turing on the topic of artificial intelligence [1], the central question "Can machines think?" began to excite interest in building systems that learn and think like people. That is a fascinating dream! Recently, the interest has been renewed again because of impressive progress with deep learning [2], in spite of great difficulties such as the Character Challenge and the Frostbite Challenge [3], to perform a variety of tasks as rapidly and flexibly as people do.

What does it mean for a system to learn and think like a person? Lake et al. argued that this system should build causal models of the world, ground learning in intuitive theories of physics and psychology, and harness compositionality and learning-to-learn [3]. They claimed that these key ideas of core ingredients would play an active and important role in producing human-like learning and thought. Undoubtedly, their claim is attractive and promising for the ultimate dream of implementing machines with human-level general intelligence. However, the claim says little about a person's ability to communicate and think in natural language, which is clearly vital for human intelligence [4]. The question is, how to develop a capacity of language for machines?

Language is a basic tool in human society, playing an essential role in communication and thought. People are accustomed to thinking in sound language (sound thinking). In different countries, people generally speak different languages. There are about 5000~7000 languages spoken all over the world, 90% of them used by less than 100000 people. As estimated by UNESCO (The United Nations' Educational, Scientific and Cultural Organization), the most widely spoken languages are: Mandarin Chinese, English, Spanish, Hindi, Arabic, Bengali, Russian, Portuguese, Japanese, German and French. In practice, a language usually takes forms of speech and text, but can be encoded into whistle, sign or braille. This may lead to an interesting question, can machines think in language with other forms, e.g. radio?

To all appearance, a Chinese can think in Chinese, an American can think in English, a Spanish can think in Spanish, and so on. From the viewpoint of daily life, all these forms of language, even including other forms such as whistle, sign and braille, should be equivalent for people to think about the world. In my opinion, this quite common point can be generalized as a first principle to establish a theory of mind (and intelligence more broadly), termed **the principle of language's relativity**, or "**the principle of symbolic relativity**" [5]. The principle is described as follows.

*All admissible forms of language are equivalent for an intelligent system to think about the world.*

In the principle of language's relativity, an admissible form means that the system can use it for thinking, i.e. the formulation of thoughts about the world. Therefore, the principle can be stated in other words,

*All admissible forms of language are equivalent with respect to the formulation of thoughts about the world.*

Note that this principle is named with inspiration from the principle of relativity in physics [6], namely,

*All admissible frames of reference are equivalent with respect to the formulation of the fundamental laws of physics.*

That is, physic laws are the same in all reference frames - inertial or non-inertial. By analogy, it can be stated that, thoughts about the world are the same in all language forms – speech, text, whistle, sign or braille. Therefore, in this sense a language form can also be regarded as a reference frame to formulate thoughts.

If taken as the first postulate of intelligence theory, the principle of language's relativity implies that language is independent of modality. This explains why any human language can be encoded into a lot of different media such as using auditory, visual, and tactile stimuli. Moreover, it can give profound and original insights to guide engineering future generations of intelligent machines. For example, principally robots can think in radio language. Such robots would be tremendously useful in space exploration, where radio language is much more convenient for them to talk with each other than sound language for lack of air. Since no person have an inborn ability to receive and send radio waves, the radio form of language is not admissible for human. Thus, radio language is a novel and creative idea for robots to think, although radio itself is certainly very ordinary for information transmission and remote control. Clearly, thinking in radio language (radio thinking), is a different way to implement intelligence than people can. One may argue that, even without language, artificial intelligence (e.g. by residual network [7], deep Q-network [8] and AlphaGo [9]) could equal or even beat human intelligence in deep learning performance of some tasks such as object recognition, video games and board games. In practical realization, it is still reasonable to require that an intelligent machine is able to communicate through language. It goes without saying, language is an essential ability for general intelligence.

Yet nobody is quite sure of what intelligence is. Perhaps in most general purpose, intelligence measures an agent's ability to achieve goals in a wide range of environments [10]. Nonetheless, this informal definition, together with the mathematic description [10], plays a limited role in design of intelligent machines, albeit bringing together some key features from many expert definitions of human intelligence. To understand the nature of intelligence, not only a far-reaching definition is further expected, but also a comprehensive theory.

What does this theory look like? At the very least, it should contain just a few first principles at the system level. The few principles must be fundamental and independent in all phenomena of intelligence, and cannot be deduced from any other principles in physics, chemistry and biology. Although these principles may not lead to anything like Maxwell equations or $E=mc^2$, they should capture the essential characteristics of intelligence comprehensively in perspectives between science and philosophy. More importantly, they should be able to make a guide to engineering intelligent machines, especially with some different intelligence from human. "Human-like" implicates imitation before grasping the nature of intelligence clearly, while "human-different" implicates creation after understanding genuinely.

One of such principles is the principle of language's relativity, others to be certain. Obviously, the principle is independent of physics, chemistry and biology. Furthermore, it can account for modality-independence of language, and give rise to a revolutionary idea of radio thinking. As a high-level intelligence envisaged for future robot astronauts, the importance of radio thinking should be emphasized again. It may overturn a public view of how robonauts talk with each other in space exploration. Imagine two robonauts are talking about the earth on the moon (see Figure 1). Traditionally, people think that they would talk in sound language, just as could be seen in some science fiction films. Nevertheless, in reality they cannot do so at all without air. Alternatively, they can talk with each other in a Morse-code radio language that lacks any neural mechanisms. This is an excellent thought experiment to show that, although human language is an ability developed on the basis of neural mechanisms in the brain, intelligent machines may have a capacity of radio language not based on these mechanisms. Thereby, high-level intelligence may not be brain-like. The brain-like intelligence tries to achieve intelligence as demonstrated by brains [11], preferably those of highly evolved creatures. But the nature of intelligence can be understood in a brain-different way, just like the secret of how to fly. Indeed, without flapping its wings, an airplane can fly in a bird-different way based on the theory of aerodynamics, not the bird's brain.

For space exploration, autonomous robonauts would be extremely helpful on the moon or the other planets. Since the environments change, there are a lot of barriers to implement these robonauts. But they should require certain kinds of human-different intelligence for high-level autonomy in actions. For example, radio thinking helps them to plan, and radio talking helps them to collaborate. Kirobo is the world's first talking robot sent into space [12]. However, Kirobo was tasked to be a companion, not an explorer. It could talk in sound language only inside the spacecraft. Radio talking may be an exceptional solution outside.

It is a terrific endeavor to engineer machines with general intelligence. A theory of intelligence requires a few first principles, for example, the principle of language's relativity. Where are other principles in the Universe? What are their implications in science? How do they play roles in technology? These questions are the challenges set ahead for remarkable years of discovery.